\documentclass[10pt,letterpaper]{article}
\usepackage[letterpaper,text={5.5in,9in},centering]{geometry}
\usepackage{float}
\usepackage{placeins}     %
\usepackage{booktabs}
\usepackage{graphicx}
\usepackage{microtype}
\usepackage[font=small,labelfont=bf,skip=6pt]{caption}
\usepackage{xcolor}
\usepackage[shortlabels]{enumitem}
\setlist[enumerate]{itemsep=1pt,topsep=4pt,parsep=0pt,leftmargin=2em}
\usepackage[hidelinks]{hyperref}
\usepackage{fancyhdr}

\fancypagestyle{plain}{\fancyhf{}\fancyfoot[C]{\thepage}%
  \fancyfoot[R]{\footnotesize\textcolor{gray}{Rev. 2026-10-19}}%
  }
\hypersetup{pdfsubject={Revision 2026-10-19}}

\newcommand{\papertitle}[1]{%
  \begin{center}{\LARGE\bfseries #1\par}\end{center}\vspace{10pt}}
\newcommand{\byline}[2]{%
  \begin{center}{\large #1\par}\vspace{4pt}{\normalsize #2\par}\end{center}
  \vspace{12pt}}

\renewenvironment{abstract}{%
  \vspace{4pt}\begin{center}{\bfseries\abstractname}\end{center}%
  \begin{list}{}{%
    \setlength{\leftmargin}{0.5in}\setlength{\rightmargin}{0.5in}%
    \setlength{\topsep}{2pt}\setlength{\itemsep}{0pt}\setlength{\parsep}{5pt}%
    \setlength{\listparindent}{0pt}}%
  \item\relax\small}%
  {\end{list}\vspace{8pt}}

\newcommand{\refsbreak}{\vspace{18pt}}

\begin{document}
\hypersetup{pdftitle={Bombadil: A Foundation Model for Plant Electrophysiology}}
\papertitle{Bombadil: A Foundation Model for Plant Electrophysiology}

\hypersetup{pdfauthor={James Kirk, Piotr Bednarski, Filip Twardy, Justin Mast, Lee von Kraus}}

\byline{James Kirk, Piotr Bednarski, Filip Twardy, Justin Mast, Lee von Kraus}{Electric Plant Co, electricplant.co}

\begin{abstract}

Living plants, like animals, use electrical signals to carry information about their physiological state across their body, from roots to branches. These signals can be recorded, as they are in animals, from a pair of differential extracellular electrodes inserted into signal-conducting tissue. While machine-learning models have been trained to predict specific plant stress states from these signals, each of these models has been trained for only a single task and none have been designed to transfer across tasks, species, or environments. These approaches do not allow easy scaling across the wide range of species and environments that are relevant to agriculture, forestry, and horticulture for both commercial and consumer plant care.

Here we present Bombadil, a foundation model for plant electrophysiology, pretrained without labels on over 4 million hours of electrical signals recorded from 100 plant species in ten distinct environments, including laboratories, greenhouses, commercial orchards, and natural forests. In this way, Bombadil is able to learn the fundamental features of plant electrical signals across a wide array of species and environments. Then, from those same electrical signals, Bombadil tracks changes in seven physiological indicators of plant stress, even among species and stressors with no labels in training.

\end{abstract}

\section{Introduction}

Plants generate electrical signals in response to environmental stimuli, and these signals propagate from the site of the stimulus to distant tissues, where they induce systemic physiological responses [Sukhov et al. 2019; Fromm \& Lautner 2007]. Three classes of long-distance signal are conventionally distinguished. The action potential (AP) is a fast (about one-second duration), all-or-none, self-propagating depolarization, triggered when a stimulus magnitude surpasses a threshold [Felle \& Zimmermann 2007; Sukhov et al. 2019]. The variation potential (VP) is a slower, graded, longer-duration response, usually occurring after wounding and other damaging stimuli [Stahlberg et al. 2006]. The system potential (SP) is a propagating, transient hyperpolarization [Zimmermann et al. 2009; Sukhov et al. 2019]. The ``plant electrome'' is the sum of these signals and a plant's other electrical activity over time [de Toledo et al. 2019]. Because the electrome can be recorded with minimally invasive electrodes from intact plants, it can provide a continuous, in situ ``readout'' of a plant's physiological state. A continuous readout based on electrical signals has advantages over existing methods of assessing plant stress states, which are costly to buy or operate, do not measure continuously, or measure from a plant's environment rather than the plant itself.

There are challenges in mapping an electrical waveform to an organism's complex physiological state. For instance, plant electrical signals mix stimulus-specific components with nonspecific ones [Mudrilov et al. 2021], and the signals are weak and easily corrupted by electrical noise [Tran et al. 2019]. They are also non-stationary: a plant primed by earlier stress responds differently to the same stimulus later [Pissolato et al. 2024; Oberkofler et al. 2021]. Finally, the overall form of the signal depends on the species [Sukhov et al. 2019], and a representation fit to one species, site, or stressor often fails to transfer to another. For these reasons, decoding these complex signals has only recently become practical, via machine learning.

Machine learning classifiers have been trained, using only plant electrical signals as input, to detect general stress [Najdenovska et al. 2021], water stress [Tran et al. 2019], nutrient deficiencies [Tran et al. 2024], nitrogen deficit [Gonz\'{a}lez i Jucl\`{a} et al. 2023], and specific external stimuli [Chatterjee et al. 2015; de Toledo et al. 2024], using methods from hand-engineered statistical features to deep learning [Qi et al. 2024]. Such models have also been found to work outside of electrically shielded labs, in commercial greenhouse conditions [Tran et al. 2019]. However, every one of these models has been focused on a single task, usually trained on one species in one setting. A new factor, species, or condition then requires a new labeled dataset and a new model. No model has yet been designed to produce a representation of plant state that transfers across tasks, species, and environments, to be deployable across a wide range of settings in agriculture, forestry, and horticulture for both commercial and consumer plant care.

To solve this generalizability issue, we take inspiration from the field of neuroscience where foundation models [Bommasani et al. 2021] have been successfully applied with the same goal. There, models train on large corpora of electrical recordings and/or fMRI scans from human or animal brains, and the general representations they learn transfer across subjects, recording hardware, and tasks [Kostas et al. 2021; Jiang et al. 2024; Qiu et al. 2025]. Electrical recordings from plants share many similarities with signals recorded from animals and humans, facilitating translation of these neuroscience techniques to the plant kingdom, a translation that has not yet been made [Wahid et al. 2026].

Electrical signals are more practical and scalable to record and use in plants than in humans or animals [Lee \& Calvo 2023]. Connecting recording electrodes to plants is straightforward, requiring no complex surgical training or facilities, and because a plant stays in one place, hundreds of individual plants can be recorded simultaneously and continuously over many months or years. Furthermore, plant signals are slow, unfolding over seconds to hours rather than milliseconds [Sukhov et al. 2019], so low-sampling-rate hardware can capture their structure and days of recording fit in a single context window. Every recording in this paper comes from one of two devices we developed in-house at Electric Plant Co: the ``Bud'' for outdoor sites and the ``Stem'' for laboratory and greenhouse settings (\S{}2.3).

In this work we present the \textbf{Bombadil Foundation Model}, a Transformer trained with a self-supervised objective, learning the structure of plant electrical signals without task labels. Because pretraining spans many species, the model is able to generalize to species held out of that pretraining. The foundation model's backbone is shared across tasks, and task-specific heads read from the representation the backbone produces. A new task then requires only a new head, not a new model built and trained from scratch.

\begin{figure}[H]
\centering
\includegraphics{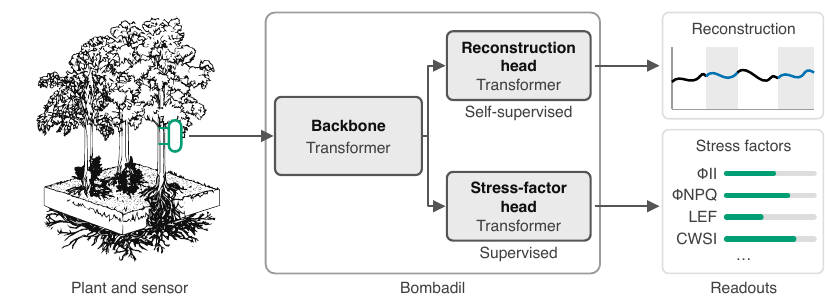}
\caption{System overview. A pair of electrodes on a plant records the electrical signal, and one Transformer backbone learns to represent it. Two task-specific heads read that representation: the patch-reconstruction head, trained without labels during pretraining, and the stress-factor head, trained on paired ground-truth readings.}
\end{figure}

Building on the foundation model, the stress-factor head we report here estimates a plant's ``vital signs,'' which are seven factors related to its photosystem and transpiration functions: five assessed by leaf chlorophyll fluorescence [Kramer et al. 2004; Baker 2008] and two assessed by leaf thermal characteristics [Jackson et al. 1981; Idso et al. 1981].

\begin{table}[H]
\centering
\small
\caption{The seven stress factors}
\begin{tabular}{lll}
\toprule
Factor & Group & Definition \\
\midrule
$\Phi$\textsubscript{II} & Fluorescence & Operating efficiency of photosystem II (PSII) \\
$\Phi$\textsubscript{NPQ} & Fluorescence & Light dissipated by regulated non-photochemical processes \\
$\Phi$\textsubscript{NO} & Fluorescence & Light dissipated by unregulated non-photochemical processes \\
Fv$'$/Fm$'$ & Fluorescence & Maximum photochemical efficiency of PSII (light-adapted) \\
LEF & Fluorescence & Rate of linear electron transport through PSII \\
LTD & Thermal & Leaf temperature differential (leaf vs. air) \\
CWSI & Thermal & Crop Water Stress Index \\
\bottomrule
\end{tabular}
\end{table}

All of these vital signs are affected by stressors, and are particularly responsive to restricted irrigation (water stress). We chose restricted irrigation as the stressor in a majority of our data collection because water stress is one of the most serious and common issues in plant care. We also tested the model's performance on data collected from plants undergoing increased salinity and glyphosate toxicity to observe the model's ability to generalize across stressors.

This paper makes six claims.

\begin{enumerate}

\item The foundation model captures the structure of the plant electrome, filling in masked spans of a signal and forecasting how it continues, even for species held out of pretraining (\S{}3.1).

\item A head on the foundation model's representation tracks stress factors within a plant under water stress (\S{}3.2).

\item The foundation model's advantage grows with the breadth of the population (\S{}3.3).

\item The estimated stress factors move as expected under stressors that are not drought, shown here for salt and glyphosate (\S{}3.4).

\item A new species can be read without labels, and calibration improves every stress factor (\S{}3.5).

\item A second sensor on a plant raises the performance floor without raising the ceiling (\S{}3.6).

\end{enumerate}

\section{Methods}

\subsection{Model architecture}

The Bombadil model is a Transformer [Vaswani et al. 2017] with a shared backbone and task-specific heads. Continuous recordings of electrical signals spanning multiple days are split into non-overlapping patches of 16 minutes, so 90 patches cover a day. Each patch holds the 16 per-minute summaries of that span (\S{}2.3) and the patch serves as one of the model's tokens, following the patch-tokenization used for time-series Transformers [Nie et al. 2023]. The backbone's Transformer layers read a window of these tokens and produce a representation of them to be shared across tasks.

Both task-specific heads also contain Transformer layers, and each reads the backbone's representation through query tokens. The patch-reconstruction head learns to fill in masked patches of the input signal. One query token stands for each masked patch, attending only to the patches that remain visible. A multilayer perceptron (MLP) projects the decoded tokens back into the shape of the input patches, along with a predicted error distribution.

Similarly, the stress-factor head learns from the labeled corpus to estimate the seven stress factors, and uses one query token per factor that it estimates. An MLP projects the decoded tokens into both a mean predicted value for each factor and a predicted error distribution around the mean.

\begin{figure}[H]
\centering
\includegraphics{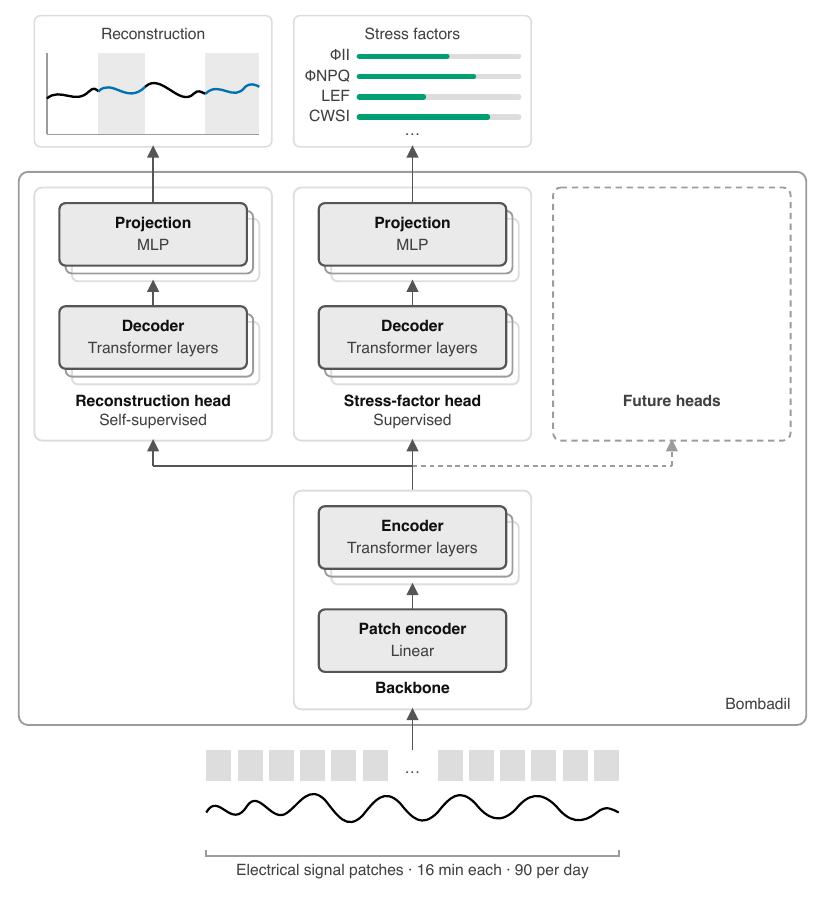}
\caption{Multi-head architecture schematic. A shared Transformer backbone over signal patches feeds task-specific heads: the patch-reconstruction head used in pretraining and the stress-factor head, with room for further heads.}
\end{figure}

\subsection{Data}

We recorded electrical signals from plants during various data generation projects (DGPs). Species, procedures, and sites varied between DGPs, and some DGPs actively applied treatments to cohorts of plants while other DGPs collected data passively. The data from all of these DGPs form our corpora.

Bombadil draws on two corpora of electrical recordings that differ in scale and in whether they carry ground-truth labels. We measure the size of these corpora in ``electrode-hours'' (one hour of recording from one sensor).

The pretraining corpus is large: over 4 million electrode-hours of raw electrical signal from more than 1,400 individual plants across 100 species, recorded at ten sites between November 2024 and July 2026. Those sites include laboratories, greenhouses, commercial orchards, and natural forests.

\begin{figure}[H]
\centering
\includegraphics{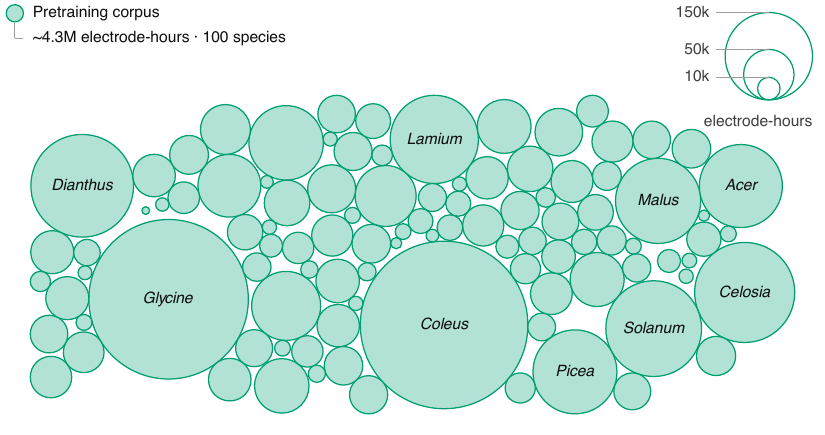}
\caption{The pretraining corpus by species. Each circle is one species, its area proportional to that species' recorded electrode-hours.}
\end{figure}

The labeled corpus is a subset within the pretraining corpus: roughly 1 million electrode-hours from over 400 individual plants across 35 species. These electrode-hours are paired with over 41,000 MultispeQ readings of the seven stress factors, which serve as ground truth when training the stress-factor head.

\begin{figure}[H]
\centering
\includegraphics{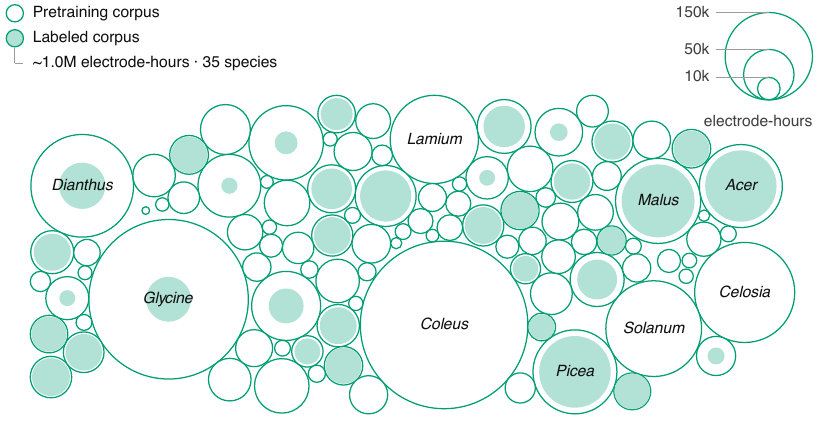}
\caption{The labeled corpus inside the pretraining corpus. Each hollow circle is one species' electrode-hours of pretraining recording, and the filled circle within it is the share of those hours that carry ground-truth labels. Both circles are drawn to the same area scale as in the pretraining corpus figure.}
\end{figure}

This two-corpus design suits a self-supervised foundation model. Representation learning runs on the large pretraining corpus without any labels, and only the labeled corpus trains the stress-factor head. The stress-factor head therefore inherits structure learned from more than 4 times as many electrode-hours and 2.9 times as many species as its own labeled data.

\subsection{Electrical signal recording devices}

Both corpora are recorded with two device types developed in-house at Electric Plant Co: the Bud and the Stem. The Bud is a low-cost, low-power outdoor device: it samples one electrode pair at 1 Hz, sends data over LoRaWAN, and runs on battery. The Stem is a device for laboratory and greenhouse settings: it samples eight electrode pairs at 120 Hz, connects over Wi-Fi, and draws power through USB.

We insert electrodes deep enough to penetrate xylem and phloem (channels that carry electrical signals in plants), so insertion depth varies by species. We use 28-gauge wire (Ag/AgCl or stainless steel) for fleshy-stemmed plants and screws (stainless steel) for woody-stemmed plants.

Each device takes point-in-time readings of the voltage across each electrode pair. The individual readings are then aggregated, either on-device or in a data pipeline, into a compressed representation with a one-minute span. These representations hold summary statistics of the minute's voltage readings, such as the mean, median, minimum, maximum, slope, skewness, and kurtosis. Aggregating this way reduces the volume of data that a device must transmit, and provides a standardized feature representation of each minute for the model's input and reconstruction output. The systemic effects of stress on the electrome develop over hours and days, so one-minute summaries capture them. Faster events such as action potentials register in the summary statistics, though the summary obscures their shape and duration.

Wireless transmission can drop data. The foundation model tolerates dropped data by design, as masked-patch pretraining (\S{}2.6.1) mimics missing spans while the model learns.

\subsection{Ground-truth labeling}

Ground truth for the stress factors comes from MultispeQ readings [Kuhlgert et al. 2016]. The MultispeQ (produced by PhotosynQ) is a handheld fluorometer, chlorophyll meter, and absorbance-based spectrometer. We take MultispeQ readings one to three times a day, depending on the DGP, and we separately record the onset of every stress treatment. Protocols vary across DGPs, but each stress DGP records treated and control cohorts of the same species simultaneously. For example, in the water-stress DGPs that train the stress-factor head, the droughted and well-watered plants are side by side.

The five fluorescence factors ($\Phi$\textsubscript{II}, $\Phi$\textsubscript{NPQ}, $\Phi$\textsubscript{NO}, Fv$'$/Fm$'$, LEF) follow standard definitions [Kramer et al. 2004; Baker 2008], and the two thermal factors (LTD, CWSI) are derived from the MultispeQ's leaf temperature, air temperature, and relative humidity readings.

For every factor except CWSI, we z-score the labels per species and train the head to regress in this z-space. CWSI needs no such step: it is derived from LTD and vapor pressure deficit within species-specific envelopes, which serve to normalize CWSI values across species. Normalizing the labels factors out species-level scale and offset, so the model is trained and evaluated on within-species variation, on a range shared across species.

The seven stress factors are not mutually independent. Four relationships link them.

\begin{enumerate}

\item The three yield terms, $\Phi$\textsubscript{II}, $\Phi$\textsubscript{NPQ}, and $\Phi$\textsubscript{NO}, partition absorbed light and sum to one [Kramer et al. 2004].

\item $\Phi$\textsubscript{II} is the product of Fv$'$/Fm$'$ and photochemical quenching.

\item LEF scales $\Phi$\textsubscript{II} by incident light.

\item CWSI is derived from LTD and vapor pressure deficit.

\end{enumerate}

We therefore report model performance per factor rather than averaging across them, since an average would weight their correlated errors unevenly.

\subsubsection{Water stress physiology}

The physiological response to water deficit varies by species and has no discrete onset. Isohydric plants close their stomata to hold leaf water potential steady, while anisohydric plants keep theirs open and let that potential fall [Sade et al. 2012]. Water deficit also develops gradually, unlike the wounding and threshold-crossing stimuli that trigger the three long-distance electrical signal classes of plants.

Both water-management strategies impair photosynthesis as water deficit deepens. Less of the absorbed light is used photochemically, so $\Phi$\textsubscript{II}, Fv$'$/Fm$'$, and LEF all fall. Since the three yield terms sum to one, $\Phi$\textsubscript{NPQ} and $\Phi$\textsubscript{NO} together take up whatever $\Phi$\textsubscript{II} no longer uses [Kramer et al. 2004]. How $\Phi$\textsubscript{NPQ} and $\Phi$\textsubscript{NO} change depends on the severity of the water deficit. When plants close their stomata, transpirational cooling drops, so the leaf warms relative to the air and both LTD and CWSI rise.

The gradual development of water stress and the range of plant water-management strategies shape how we prepare the ground-truth labels. The stress factors that respond to drought change continuously, with no natural discrete boundary that would allow drought to be treated as a binary state across species. Differing strategies between species also mean the same deficit produces different amounts of movement in the stress factors. That difference in response scale is one of the species-level differences that per-species z-scoring normalizes (\S{}2.4).

\subsubsection{Leaf-to-leaf variation}

A MultispeQ reading is taken on one leaf, and leaves on the same plant differ [Bielczynski et al. 2017; Sperdouli et al. 2021]. We tag one leaf per plant for MultispeQ readings and return to it for every reading, so any leaf-specific offset in ground truth stays with the plant for the length of the DGP.

\subsection{Splits and ablations}

The default holdout is at the plant level. Plants are split 60/20/20 into training, validation, and test sets. Each plant keeps that assignment for both training phases, meaning a plant in the foundation model's test set is also in the stress-factor head's test set.

Two species-ablation experiments add five-fold cross-validation grouped by species: we partition the species into five groups and withhold each species group in turn from training. Ablation statistics come from each ablation's withheld group intersected with the default test split, aligning the plants scored under both. The first experiment withholds species from the foundation model's pretraining (\S{}3.1). The second experiment withholds species from the stress-factor head's labeled training set (\S{}3.5).

\subsection{Foundation model training and evaluation}

\subsubsection{Pretraining}

We pretrain the foundation model using masked patch modeling (MPM) on the large pretraining corpus (\S{}2.2). During MPM, we hide a fraction of the input patches and train the model to reconstruct them from the visible patches through the patch-reconstruction head. Masked-patch pretraining of this kind underlies language [Devlin et al. 2019] and vision [He et al. 2022] models, and masked-value reconstruction has been used for multivariate time series [Zerveas et al. 2021]. Closest to our setting, masked-patch pretraining has been applied to brain electrophysiology [Jiang et al. 2024].

\begin{figure}[H]
\centering
\includegraphics{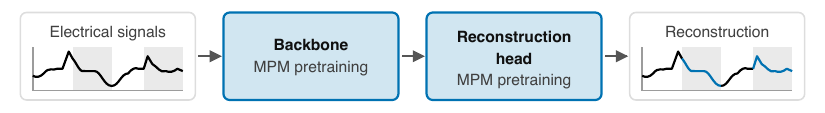}
\caption{Masked-patch pretraining. A fraction of input patches are hidden. The backbone and patch-reconstruction head train together to reconstruct them from the patches that remain visible. Grey marks the hidden spans, and the reconstruction is drawn across them against the observed signal.}
\end{figure}

MPM trains on four-day windows with 360 patches each, one token per patch. The model sees over 175 billion tokens in pretraining. We hide 60 percent of input patches randomly at each step. MPM pretraining of the foundation model takes approximately 84 hours on four NVIDIA A100 GPUs.

\subsubsection{Metrics}

For the foundation model, we report reconstruction and forecasting error on held-out recordings (\S{}3.1). Both are RMSE-w$\sigma$, a root-mean-square error in units of each window's own standard deviation over the full four days. The model reconstructs every channel of the per-minute representation (\S{}2.3), and RMSE-w$\sigma$ is computed per channel and averaged across channels. Recorded voltages span orders of magnitude between plants and sensors, so normalizing within each window is what makes errors comparable across windows. An error of 1.0 on RMSE-w$\sigma$ means the model missed by as much as the window itself varies.

\begin{figure}[H]
\centering
\includegraphics{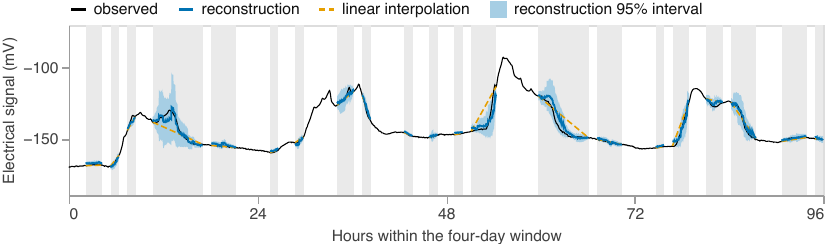}
\caption{Masked patch modeling on one four-day window from a held-out \emph{Glycine max} plant, one sensor. Grey marks the patches hidden from the model at evaluation, half the window in 64-minute blocks (\S{}2.6.4). The observed signal runs the whole window, and the model's reconstruction covers the hidden spans with the estimated 95 percent interval shaded. The dashed line is the linear interpolation reference predictor.}
\end{figure}

Forecasting is not a pretraining objective, but the patch-reconstruction head forecasts as well. Forecasting is the same operation as MPM and differs only in the mask, which covers the end of the window rather than patches scattered throughout (\S{}2.6.4). Forecasting is central to time-series foundation models [Das et al. 2024; Ansari et al. 2024], so measuring model performance in forecasting serves as a useful evaluation of the model's understanding of time-series structure. We report forecast error over the whole masked day as well as by lead time within the masked day.

\begin{figure}[H]
\centering
\includegraphics{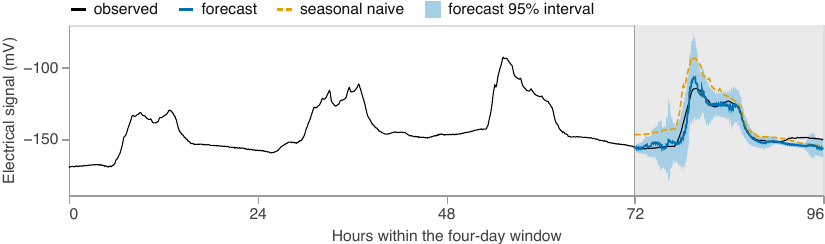}
\caption{The same window under forecasting. Grey marks the final day, hidden and forecasted from the three days before it. The forecast is drawn against the observed signal with its 95 percent interval shaded, and the dashed line is the seasonal naive reference predictor.}
\end{figure}

\subsubsection{Reference predictors}

Model-free reference predictors set a scale that RMSE-w$\sigma$ lacks. We score them through the same evaluations, on the same windows.

Reconstruction is scored against the linear interpolation reference. Forecasting has three references: seasonal naive, persistence, and the window mean.

\subsubsection{Masking schemes}

During MPM evaluation (\S{}3.1) we mask differently than during training. Evaluation hides half the window in 64-minute blocks of four patches each, chosen at random. These masks are wider than the model's patch width so adjacent patches are often both hidden and the gaps vary in length.

The forecasting evaluations (\S{}3.1) come from the same model on the same four-day windows as MPM evaluations, but with the final day fully masked.

\subsection{Stress-factor head training and evaluation}

\subsubsection{Training}

The stress-factor head is trained on the smaller labeled corpus, using the paired ground-truth readings as labels (\S{}2.4). The head sees a three-day window of 270 patches, encoded by the foundation model's backbone, and estimates each factor's state at the end of the window. The backbone is frozen during head training, with the exception of the ``task-trained backbone'' outlined in \S{}2.8.

\begin{figure}[H]
\centering
\includegraphics{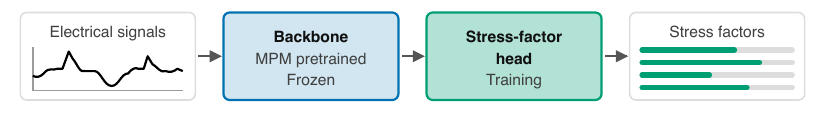}
\caption{Stress-factor head training. The backbone has been pretrained with MPM and is frozen. The stress-factor head trains on the paired ground-truth readings of the labeled corpus (\S{}2.4). The head reads a window of the electrical signal and estimates every stress factor's state at the end of the window.}
\end{figure}

The head is trained with two losses combined: the concordance correlation coefficient [Lin 1989] fits the mean estimate, and a negative log-likelihood on the full distribution fits both the mean and the variance estimates.

Stress-factor head training takes approximately 40 hours on one NVIDIA L4 GPU.

\subsubsection{Metrics}

We report four measures per factor: trajectory r, concordance correlation coefficient (CCC), RMSE-s$\sigma$, and area under the receiver operating characteristic curve (AUC). Statistics are computed per sensor, meaning one pair of electrodes on one plant, and we report sensor and plant counts rather than point counts, since readings from paired sensors on the same plant are not independent.

A plant's stress factors rise and fall over time, and we refer to that as ``movement'' of the plant's paired ground-truth measurements. Trajectory r measures how closely the head tracks movement, and is calculated as the Pearson correlation between estimates and measurements over the weeks a plant is recorded. Trajectory r is calculated for each sensor and averaged across sensors in Fisher-z space, with every sensor counting equally.

The head's other three measures ask a different question, which we call ``position'': where a plant sits against other plants on each factor's shared scale. CCC is a correlation with a penalty for systematic bias against the 1:1 line [Lin 1989], so CCC measures movement and position together. RMSE-s$\sigma$ is a root-mean-square error in units of each factor's standard deviation within a species, so the seven stress factors' errors are comparable. AUC treats each estimate as a prediction of whether the plant currently sits above or below the median for that factor.

We anchor these four measures against a no-skill baseline: the score a model would earn by predicting each species' mean and nothing more. Such a predictor has no correlation with the measurements, so trajectory r and CCC are 0. The predictor's error equals the species' own standard deviation, which is 1.000 on the RMSE-s$\sigma$ scale. A constant estimate cannot separate plants above the median from those below, giving an AUC of 0.500.

\subsubsection{Evaluation populations}

In \S{}3.3, we evaluate on all 35 labeled species. Every other result that uses ground-truth readings works on the six densely labeled species.

Two requirements produce the densely labeled subset: trajectory r needs a long record from one plant before there is movement to correlate, and the per-species evaluations need several plants behind each metric. The other three measures (CCC, RMSE-s$\sigma$, AUC) pool plants together, so a short record still contributes to those metrics. Six of the 35 species clear both criteria to be considered densely labeled.

The seven stress factors do not all cover the same population. $\Phi$\textsubscript{II}, $\Phi$\textsubscript{NPQ}, $\Phi$\textsubscript{NO}, and Fv$'$/Fm$'$ cover all six densely labeled species, 35 plants, and 70 sensors. LEF covers five species, 29 plants, and 58 sensors, because the MultispeQ returned no LEF for \emph{Dianthus barbatus}. LTD and CWSI cover four species, 24 plants, and 48 sensors, because the MultispeQ's leaf temperature readings were not viable for \emph{Phoenix roebelenii} or \emph{Picea abies} (\S{}4.4.1).

\subsubsection{Error structure}

A sensor's estimates can sit consistently high or low against its labels, and that constant difference is its offset. We partition each factor's pooled squared error into two parts: the squared offset of each sensor, and the remainder of the error once the offset is removed. The two sum exactly to the pooled error. The offset does not affect trajectory r, which is computed with each sensor's own mean removed. On the other hand, CCC, RMSE-s$\sigma$, and AUC carry the offset in full.

Repeat MultispeQ readings on one plant in a single session disagree with each other, so no model could follow these labels perfectly. That disagreement sets a ceiling on trajectory r. We estimate the ceiling from up to 177 plants and 9,376 multi-reading sessions, pooled across all splits because label noise belongs to the instrument rather than to a split. Those repeat readings were gathered across different leaves, so their disagreement carries leaf-to-leaf variation on top of the instrument's own noise.

The head's labels come from one tagged leaf per plant (\S{}2.4.2), where leaf-to-leaf variation is a fixed offset rather than per-reading noise. We hold the leaf contribution constant when estimating the ceiling, since a fixed offset does not affect trajectory r. Excluding leaf-to-leaf differences puts the ceiling higher and attributes less of the head's shortfall to label noise, which is the conservative direction.

The ceiling runs 0.91 to 0.97 on the fluorescence factors and 0.75 on LTD, and covers movement only. CWSI has no ceiling, being derived from other readings rather than measured directly.

\subsection{Backbone comparisons}

The stress-factor head reads the representation produced by the backbone rather than raw electrical signal, so the quality of the backbone limits what the head can recover. To measure what pretraining adds, we compare three backbones in \S{}3.3.

The frozen random backbone is randomly initialized and stays frozen, so the head reads what an untrained network produces. The task-trained backbone also starts from random initialization but trains alongside the head on the water-stress labels, so its representation is learned entirely within the task. The foundation model is pretrained without labels and frozen. The foundation model and the frozen random backbone differ only in pretraining, so the gap between them isolates the quality of the pretrained representation. The backbone architecture, the head architecture, and the labeled corpus the head trains on are identical in all three.

\begin{figure}[H]
\centering
\includegraphics{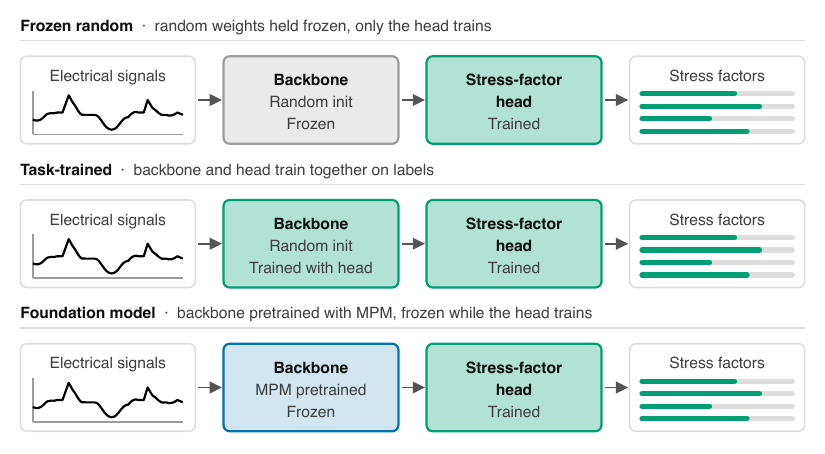}
\caption{The three backbone training schemes. Each row is one complete pipeline, running left to right from the electrical signal to the estimated stress factors. The rows differ only in the backbone, and each backbone is labeled with what it starts from and whether it trains.}
\end{figure}

The frozen random backbone is the baseline for both comparisons, so the other two are directly comparable. The frozen random backbone scores better than the no-skill baseline: random weights make usable features, an effect credited to convolution and pooling priors [Saxe et al. 2011], and a randomly initialized Transformer smooths the sequence it is given rather than building structure over it [Liu et al. 2026]. A head on that backbone should read a few densely labeled species and fall away across many (\S{}3.3).

The backbone comparison runs on every plant with paired ground-truth readings on all labeled species. Six densely labeled species (\S{}2.7.3) alone cannot demonstrate how far a representation reaches, but the full labeled population (\S{}2.2) can. All three backbones read the same plants, so the choice of population favors none of them.

\subsection{Non-drought stressors}

The stress-factor head's labels in training cover only drought as a stressor, so we also test how the stress-factor head behaves under other stressors. Water stress has an electrophysiology of its own [Comparini et al. 2020; R\'{i}os-Rojas et al. 2014; Gil \& Vargas 2023], and we aim for the head to estimate stress-factor labels accurately regardless of the specific stressor used. Responses are stimulus-specific, so the electrical signal differs from one stressor to the next [Mudrilov et al. 2021], and recovering which stressor relates to a given systemic signal is an open problem [Huber \& Bauerle 2016]. A head that had learned a drought-specific signature may not read the plant's physiological state, quantified by MultispeQ readings, correctly once the stressor changes.

We probe Bombadil's response to other stressors using data from a DGP that predates our use of the MultispeQ: one laboratory, one time window, and three contemporaneous cohorts of \emph{Coleus scutellarioides} receiving either increased salinity, glyphosate toxicity, or no treatment (control). Each treatment was applied at onset, after which watering without the stressor resumed. No MultispeQ readings exist for this species in any DGP, so the head has never seen a MultispeQ-labeled example of the species or of either stressor, and no ground-truth readings exist to evaluate estimates against.

Therefore, in \S{}3.4, we ask a directional question instead of a quantitative one: how do the head's estimates move when plants are under these stressors? We derive the expected directions from the literature on each stressor, and \S{}4.3 sets out those derivations and their constraints. We use Cohen's d to measure how far the treated cohorts separate from the untreated control cohort, where 0.8 is the conventional threshold for a large effect [Cohen 1988]. We also report a 95 percent interval on each effect size, computed as a percentile bootstrap over plants. The directional test covers four of the seven factors: the three yield terms ($\Phi$\textsubscript{II}, $\Phi$\textsubscript{NPQ}, $\Phi$\textsubscript{NO}) and Fv$'$/Fm$'$.

\subsection{Calibration of the stress-factor head}

We perform a species ablation study on the stress-factor head (\S{}2.5) with two arms, which differ only in whether a species appears in the head's training labels. The ``uncalibrated'' arm trains the head with the evaluated species excluded. The ``calibrated'' arm trains the head with every species present. The gap between the two arms is the gain from calibrating on a new species, measured on species whose labels we already hold (\S{}2.7.3). We report the calibration gain as the difference between the arms (\S{}3.5), with a 95 percent interval on each gain computed as a percentile bootstrap over plants.

\subsection{Paired sensors}

Every plant of the six densely labeled species (\S{}2.7.3) was recorded with two sensors simultaneously. We use these sensor pairs to measure the value of a second sensor on the same plant without a new experiment (\S{}3.6).

We compare each individual sensor against the sensors as a pair. This paired-sensor comparison uses the same plants, species, labels, and stress-factor head as \S{}3.2, without refitting. The two sensors' estimates for a plant are averaged at the prediction level into a virtual blended sensor, which is scored using the same evaluations as the individual sensors. Per-plant order statistics give the better and worse sensor of each pair, ranked by trajectory r.

Averaging can only remove error that differs between the two sensors, so we measure how much per-sensor offset (\S{}2.7.4) the two sensors share. We correlate the two sensors' per-plant offsets across plants, then compare the variance of the virtual blended sensor's offset against the variance of an individual sensor's offset.

\section{Results}

\subsection{The foundation model captures the structure of the plant electrome}

Before adding a physiological head, we ask what the foundation model has learned about electrical signals on its own.

We evaluate two kinds of model: the foundation model's backbone pretrained on every species, and five ablation models, each trained with one group of species held out (\S{}2.5). Both are scored on the same held-out population: 285 plants, 469 sensors, and 80 species. The foundation model reconstructs hidden patches at 0.343 RMSE-w$\sigma$. A straight line drawn across each gap reaches 0.390, so the foundation model removes 12 percent of that reference's mean RMSE-w$\sigma$.

\begin{table}[H]
\centering
\small
\caption{Reconstruction error over the hidden patches, against linear interpolation}
\begin{tabular}{llr}
\toprule
Predictor & What it predicts & RMSE-w$\sigma$ \\
\midrule
Foundation model &  & \phantom{$-$}\textbf{0.343} \\
Foundation model, species held out &  & \phantom{$-$}0.361 \\
Linear interpolation & a line across each gap & \phantom{$-$}0.390 \\
\bottomrule
\end{tabular}
\end{table}

The same model, trained only to fill masked spans, also forecasts the masked final day, reaching 0.912 RMSE-w$\sigma$ and removing 10 percent of the mean RMSE-w$\sigma$ of seasonal naive, the strongest of the three reference predictors.

\begin{table}[H]
\centering
\small
\caption{Forecast error over the masked day, against reference predictors}
\begin{tabular}{llr}
\toprule
Predictor & What it predicts & RMSE-w$\sigma$ \\
\midrule
Foundation model &  & \phantom{$-$}\textbf{0.912} \\
Foundation model, species held out &  & \phantom{$-$}0.968 \\
Seasonal naive & the value 24 hours earlier & \phantom{$-$}1.011 \\
Persistence & the last value in the context & \phantom{$-$}1.085 \\
Window mean & the mean of the context & \phantom{$-$}1.126 \\
\bottomrule
\end{tabular}
\end{table}

A single RMSE-w$\sigma$ for the whole day hides a decay. In the first hour the foundation model's error is a quarter of seasonal naive's, and over the final twelve hours the foundation model and seasonal naive are within two percent of each other. Persistence shows the same decay from the other direction. Persistence matches the foundation model in the first hour, then falls behind seasonal naive by the twelve-hour mark. Although the model sees only the signal's past, it is predicting a day ahead for a plant whose state depends on weather, watering, and handling that the model cannot observe. Neither seasonal naive nor persistence leads across the whole day, and the foundation model beats both at every horizon.

\begin{table}[H]
\centering
\small
\caption{Forecast RMSE-w$\sigma$ by lead time}
\begin{tabular}{lrrrr}
\toprule
Lead & Foundation model & Species held out & Seasonal naive & Persistence \\
\midrule
0--1h & \phantom{$-$}\textbf{0.202} & \phantom{$-$}0.279 & \phantom{$-$}0.800 & \phantom{$-$}0.210 \\
1--3h & \phantom{$-$}\textbf{0.431} & \phantom{$-$}0.529 & \phantom{$-$}0.821 & \phantom{$-$}0.483 \\
3--6h & \phantom{$-$}\textbf{0.700} & \phantom{$-$}0.775 & \phantom{$-$}0.840 & \phantom{$-$}0.764 \\
6--12h & \phantom{$-$}\textbf{0.838} & \phantom{$-$}0.893 & \phantom{$-$}0.884 & \phantom{$-$}1.068 \\
12--24h & \phantom{$-$}\textbf{0.946} & \phantom{$-$}0.980 & \phantom{$-$}0.960 & \phantom{$-$}1.132 \\
\bottomrule
\end{tabular}
\end{table}

The structure the foundation model learns is shared across species. An ablation model, which never saw a given species, still beats every reference predictor on that species over the full day. Pretraining on that species improves performance on it: reconstruction RMSE-w$\sigma$ falls from the ablation model's 0.361 to the foundation model's 0.343 on average, forecast RMSE-w$\sigma$ from 0.968 to 0.912 on average, and the foundation model removes 10 percent of seasonal naive's mean RMSE-w$\sigma$ where the ablation model removes 4 percent. The ablation model drops to second place at every horizon, beaten early by persistence and late by seasonal naive, but wins over the day because neither reference leads throughout.

\subsection{A head on the representation tracks stress factors within a plant under water stress}

Estimating the seven stress factors from electrophysiology alone requires a representation that generalizes across species' differences in stress response and onset. That is why we train the stress-factor head on top of the foundation model, which \S{}3.1 shows generalizes to species held out of pretraining. We then evaluate the head's estimates against the held-out ground-truth readings from the labeled corpus.

\begin{figure}[H]
\centering
\includegraphics{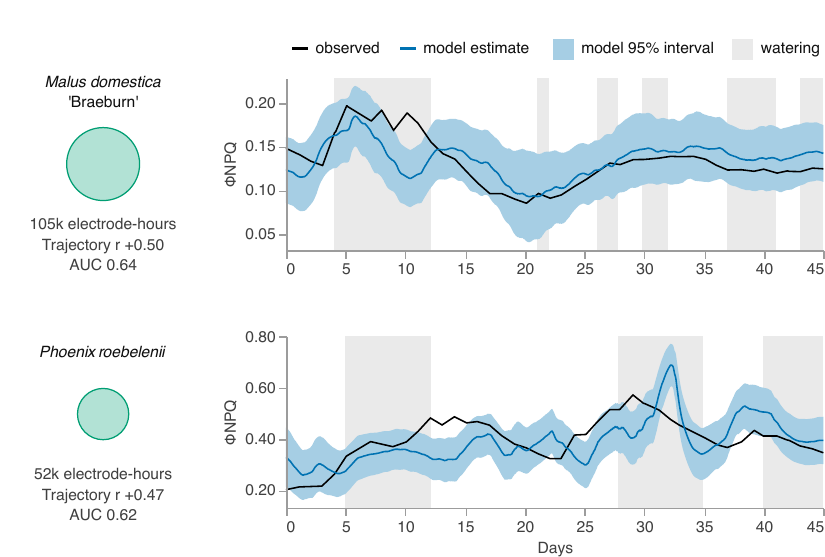}
\caption{Tracking at two corpus sizes. $\Phi$\textsubscript{NPQ} over 45 days for one droughted plant from the test split of each of two densely labeled species. The head's estimate and 95 percent interval are drawn against the observed ground-truth trace, and grey bands mark watering. Each circle is that species' labeled corpus on the same area scale as \S{}2.2's figures, and the trajectory r and AUC beside it are the species'.}
\end{figure}

\begin{table}[H]
\centering
\small
\caption{Per-factor results on the held-out test split}
\begin{tabular}{lrrrrrr}
\toprule
Factor & Plants & Sensors & Trajectory r & CCC & RMSE-s$\sigma$ & AUC \\
\midrule
\emph{No skill} &  &  & \phantom{$-$}\emph{0} & \phantom{$-$}\emph{0} & \phantom{$-$}\emph{1.000} & \phantom{$-$}\emph{0.500} \\
$\Phi$\textsubscript{II} & \phantom{$-$}35 & \phantom{$-$}70 & \phantom{$-$}0.431 & \phantom{$-$}0.293 & \phantom{$-$}1.021 & \phantom{$-$}0.697 \\
$\Phi$\textsubscript{NPQ} & \phantom{$-$}35 & \phantom{$-$}70 & \phantom{$-$}0.452 & \phantom{$-$}0.293 & \phantom{$-$}0.989 & \phantom{$-$}0.686 \\
$\Phi$\textsubscript{NO} & \phantom{$-$}35 & \phantom{$-$}70 & \phantom{$-$}0.448 & \phantom{$-$}0.393 & \phantom{$-$}0.968 & \phantom{$-$}0.714 \\
Fv$'$/Fm$'$ & \phantom{$-$}35 & \phantom{$-$}70 & \phantom{$-$}0.450 & \phantom{$-$}0.311 & \phantom{$-$}0.983 & \phantom{$-$}0.704 \\
LEF & \phantom{$-$}29 & \phantom{$-$}58 & \phantom{$-$}0.707 & \phantom{$-$}0.554 & \phantom{$-$}0.831 & \phantom{$-$}0.807 \\
LTD & \phantom{$-$}24 & \phantom{$-$}48 & \phantom{$-$}0.292 & \phantom{$-$}0.234 & \phantom{$-$}1.129 & \phantom{$-$}0.610 \\
CWSI & \phantom{$-$}24 & \phantom{$-$}48 & \phantom{$-$}0.269 & \phantom{$-$}0.198 & \phantom{$-$}1.142 & \phantom{$-$}0.615 \\
\bottomrule
\end{tabular}
\end{table}

The head tracks movement on every stress factor. Trajectory r runs from 0.269 on CWSI to 0.707 on LEF across the six densely labeled species (\S{}2.7.3). Of the six stress factors that have a label ceiling (\S{}2.7.4), $\Phi$\textsubscript{II}, $\Phi$\textsubscript{NPQ}, $\Phi$\textsubscript{NO}, and Fv$'$/Fm$'$ reach 46 to 49 percent of that ceiling, LEF reaches 73 percent, and LTD reaches 39 percent.

About half of the head's squared error is a per-sensor offset (\S{}2.7.4), running from 42 percent on LTD to 54 percent on $\Phi$\textsubscript{NPQ}. Three stress factors ($\Phi$\textsubscript{II}, LTD, and CWSI) score worse than the no-skill baseline on RMSE-s$\sigma$, and subtracting that offset would make all three better than the baseline. The head's performance on position measures is therefore driven by a fixed error on each sensor, rather than by how well it follows the plant.

\begin{table}[H]
\centering
\footnotesize
\setlength{\tabcolsep}{3pt}
\caption{Per-species within-plant tracking (trajectory r)}
\begin{tabular}{lrrrrrrrr}
\toprule
Species & Plants & $\Phi$\textsubscript{II} & $\Phi$\textsubscript{NPQ} & $\Phi$\textsubscript{NO} & Fv$'$/Fm$'$ & LEF & LTD & CWSI \\
\midrule
\emph{Acer rubrum} `October Glory' & \phantom{$-$}7 & \phantom{$-$}0.549 & \phantom{$-$}0.475 & \phantom{$-$}0.452 & \phantom{$-$}0.430 & \phantom{$-$}0.633 & \phantom{$-$}\textbf{0.480} & \phantom{$-$}\textbf{0.373} \\
\emph{Malus domestica} `Braeburn' & \phantom{$-$}7 & \phantom{$-$}\textbf{0.678} & \phantom{$-$}0.496 & \phantom{$-$}0.578 & \phantom{$-$}0.422 & \phantom{$-$}0.788 & \phantom{$-$}0.271 & \phantom{$-$}0.250 \\
\emph{Picea abies} & \phantom{$-$}7 & \phantom{$-$}0.350 & \phantom{$-$}\textbf{0.738} & \phantom{$-$}\textbf{0.794} & \phantom{$-$}\textbf{0.825} & \phantom{$-$}\textbf{0.901} &  &  \\
\emph{Dianthus barbatus} & \phantom{$-$}6 & \phantom{$-$}0.246 & \phantom{$-$}0.204 & \phantom{$-$}0.162 & \phantom{$-$}0.237 &  & \phantom{$-$}0.320 & \phantom{$-$}0.335 \\
\emph{Codiaeum variegatum} & \phantom{$-$}4 & $-$0.030 & $-$0.074 & $-$0.283 & $-$0.202 & $-$0.110 & $-$0.092 & \phantom{$-$}0.001 \\
\emph{Phoenix roebelenii} & \phantom{$-$}4 & \phantom{$-$}0.480 & \phantom{$-$}0.474 & \phantom{$-$}0.352 & \phantom{$-$}0.412 & \phantom{$-$}0.629 &  &  \\
\bottomrule
\end{tabular}
\end{table}

Performance varies by species, and it varies by stress factor within a species. \emph{Dianthus barbatus} is read better on the two thermal factors than on any of its fluorescence factors, while \emph{Malus domestica} favors the fluorescence factors, and \emph{Acer rubrum} `October Glory' is consistent across both fluorescence and thermal factors.

\begin{figure}[H]
\centering
\includegraphics{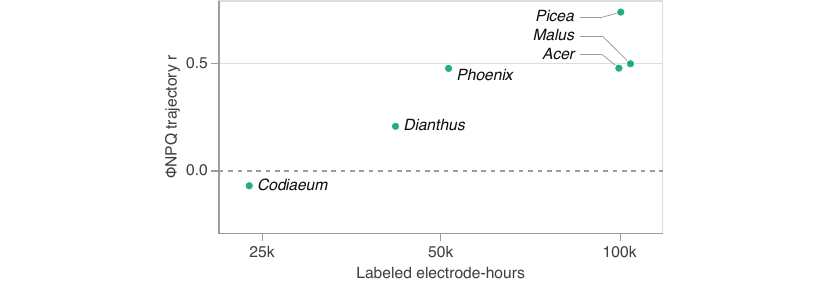}
\caption{Corpus size against tracking, species by species. $\Phi$\textsubscript{NPQ} trajectory r over the test split for each of the six densely labeled species, plotted against each species' labeled electrode-hours. The dashed line marks the no-skill value.}
\end{figure}

Corpus size may drive performance, but it does not explain all of the variance between species. For example, \emph{Malus domestica} is backed by 105,000 electrode-hours and \emph{Phoenix roebelenii} by 52,000 electrode-hours, but the two species track $\Phi$\textsubscript{NPQ} within 0.03 of each other in trajectory r. \emph{Codiaeum variegatum} holds the smallest labeled corpus out of the densely labeled species at 24,000 electrode-hours and is the worst-performing species, reading at or below zero on every stress factor.

Performance varies as much from sensor to sensor as it does from species to species. On every stress factor, the upper quartile of sensors leads the lower by at least 0.38 in trajectory r. Between 9 and 25 percent of sensors track negatively, moving against the measurement rather than failing to follow it. \S{}3.6 returns to the difference between sensors on one plant.

\subsection{The foundation model's advantage grows with the breadth of the population}

To measure what pretraining adds, and whether the answer depends on how many species the model has to learn from, we compare three backbones (\S{}2.8): a frozen random backbone, a task-trained backbone, and the foundation model.

\begin{table}[H]
\centering
\small
\caption{CCC by backbone on every labeled species}
\begin{tabular}{lrrrrr}
\toprule
Factor & Plants & Sensors & Frozen random & Task-trained & Foundation model \\
\midrule
\emph{No skill} &  &  & \phantom{$-$}\emph{0} & \phantom{$-$}\emph{0} & \phantom{$-$}\emph{0} \\
$\Phi$\textsubscript{II} & \phantom{$-$}91 & \phantom{$-$}182 & \phantom{$-$}0.092 & \phantom{$-$}0.134 & \phantom{$-$}\textbf{0.193} \\
$\Phi$\textsubscript{NPQ} & \phantom{$-$}91 & \phantom{$-$}182 & \phantom{$-$}0.083 & \phantom{$-$}0.119 & \phantom{$-$}\textbf{0.191} \\
$\Phi$\textsubscript{NO} & \phantom{$-$}91 & \phantom{$-$}182 & \phantom{$-$}0.131 & \phantom{$-$}0.192 & \phantom{$-$}\textbf{0.246} \\
Fv$'$/Fm$'$ & \phantom{$-$}91 & \phantom{$-$}182 & \phantom{$-$}0.068 & \phantom{$-$}0.114 & \phantom{$-$}\textbf{0.198} \\
LEF & \phantom{$-$}76 & \phantom{$-$}151 & \phantom{$-$}0.217 & \phantom{$-$}0.195 & \phantom{$-$}\textbf{0.283} \\
LTD & \phantom{$-$}73 & \phantom{$-$}146 & \phantom{$-$}0.042 & \phantom{$-$}0.096 & \phantom{$-$}\textbf{0.122} \\
CWSI & \phantom{$-$}72 & \phantom{$-$}144 & \phantom{$-$}0.029 & \phantom{$-$}0.087 & \phantom{$-$}\textbf{0.117} \\
\bottomrule
\end{tabular}
\end{table}

The foundation model results above sit lower than the CCC values in \S{}3.2 because this evaluation covers every labeled species, rather than only the densely labeled species (\S{}2.7.3).

Changing the backbone barely moves trajectory r, but it moves CCC substantially. Averaged over the seven stress factors, the foundation model doubles CCC over the frozen random backbone. The task-trained backbone is the representation the labeled corpus builds on its own, and it sits between the frozen random backbone and the foundation model.

\begin{table}[H]
\centering
\small
\caption{The foundation model's gain over a frozen random backbone}
\begin{tabular}{lrrrr}
\toprule
Group & Species & Plants & $\Delta$ CCC & $\Delta$ trajectory r \\
\midrule
Densely labeled species & \phantom{$-$}6 & \phantom{$-$}35 & \phantom{$-$}0.043 & \phantom{$-$}0.025 \\
Every other species & \phantom{$-$}29 & \phantom{$-$}56 & \phantom{$-$}\textbf{0.115} & \phantom{$-$}0.024 \\
\bottomrule
\end{tabular}
\end{table}

The foundation model drives improved performance both among the densely labeled species and among all the remaining species as a group. However, the gain in CCC outside the six densely labeled species is larger by a factor of 2.7, while the gain in trajectory r is unchanged between the two species groups. The frozen random backbone reaches 0.282 in CCC within the six species and collapses to 0.014 outside them. Across a wider population, the foundation model buys position, not movement. Beyond the six species, the foundation model is what lets the head put a plant's estimates at the right value on the common scale.

\subsection{The estimated stress factors move as expected under stressors that are not drought}

We evaluate the stress-factor head's ability to generalize on electrical recordings from \emph{Coleus scutellarioides} under increased salinity and glyphosate toxicity (\S{}2.9).

\begin{figure}[H]
\centering
\includegraphics{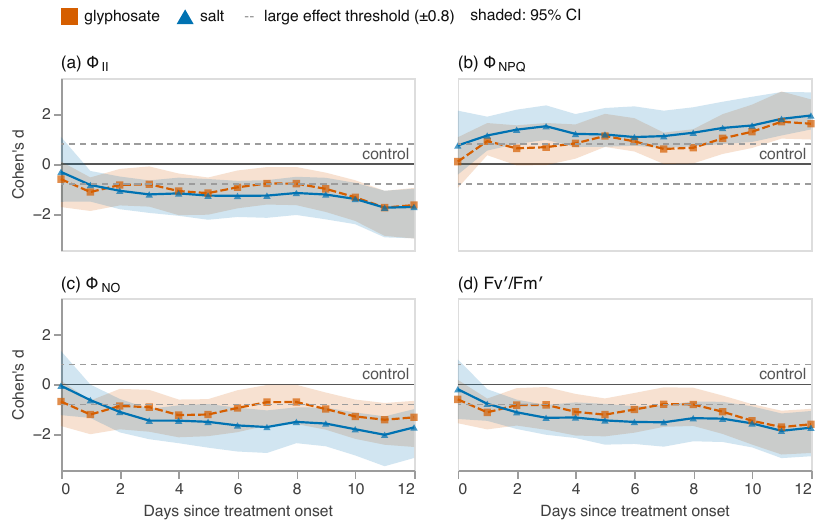}
\caption{Treatment effect against a contemporaneous control. Cohen's d for four of the head's estimated stress factors in salt-treated and glyphosate-treated \emph{Coleus scutellarioides}, against the untreated cohort, over twelve days from treatment onset. Shading is the 95 percent bootstrap interval, and the dashed lines mark the conventional large-effect threshold of $\pm$0.8.}
\end{figure}

The head's estimates move in the directions the literature expects (\S{}4.3), and the effect sizes are large (beyond $\pm$0.8). Measured against the untreated cohort, the estimated $\Phi$\textsubscript{II}, $\Phi$\textsubscript{NO}, and Fv$'$/Fm$'$ fall and the estimated $\Phi$\textsubscript{NPQ} rises under both treatments through the recorded period. Across the four stress factors and both treatments, the magnitude of Cohen's d peaks between 1.42 and 2.03. Every stress factor for which the literature gives a direction moves in the expected direction.

\subsection{A new species can be read without labels, and calibration improves every stress factor}

The foundation model generalizes to species held out of its pretraining (\S{}3.1). We ask the same of the stress-factor head, withholding species from the head's labeled training set and reading those species with a head that never saw an example of them. The calibrated result below is \S{}3.2's result, unchanged.

\begin{table}[H]
\centering
\small
\caption{Trajectory r with and without calibration}
\begin{tabular}{lrrrr}
\toprule
Factor & Uncalibrated & Calibrated & Gain & Gain 95\% CI \\
\midrule
$\Phi$\textsubscript{II} & \phantom{$-$}0.362 & \phantom{$-$}0.431 & \phantom{$-$}0.069 & $-$0.063 to 0.214 \\
$\Phi$\textsubscript{NPQ} & \phantom{$-$}0.238 & \phantom{$-$}\textbf{0.452} & \phantom{$-$}0.214 & \phantom{$-$}0.086 to 0.343 \\
$\Phi$\textsubscript{NO} & \phantom{$-$}0.183 & \phantom{$-$}\textbf{0.448} & \phantom{$-$}0.265 & \phantom{$-$}0.119 to 0.386 \\
Fv$'$/Fm$'$ & \phantom{$-$}0.178 & \phantom{$-$}\textbf{0.450} & \phantom{$-$}0.272 & \phantom{$-$}0.102 to 0.434 \\
LEF & \phantom{$-$}0.629 & \phantom{$-$}\textbf{0.707} & \phantom{$-$}0.078 & \phantom{$-$}0.010 to 0.138 \\
LTD & \phantom{$-$}0.176 & \phantom{$-$}0.292 & \phantom{$-$}0.116 & $-$0.095 to 0.285 \\
CWSI & \phantom{$-$}0.189 & \phantom{$-$}0.269 & \phantom{$-$}0.080 & $-$0.120 to 0.253 \\
\bottomrule
\end{tabular}
\end{table}

The head tracks every stress factor on a species it never trained on. The uncalibrated head reaches 19 to 65 percent of the tracking the labels permit, where the calibrated head reaches 39 to 73 percent. Calibrating on a species improves all seven stress factors, but by amounts that differ and are not all measurable at this sample size. Calibration provides 47 to 60 percent of the total calibrated result on $\Phi$\textsubscript{NPQ}, $\Phi$\textsubscript{NO}, and Fv$'$/Fm$'$, and those are the three largest gains. On LEF, calibration supplies 11 percent of the calibrated result, though it starts as the highest-performing stress factor of the seven. For $\Phi$\textsubscript{II}, LTD, and CWSI the confidence interval of the gain spans zero, so we cannot say with certainty whether calibration helps the stress-factor head's ability to track movement. At 24 to 35 plants per stress factor, that uncertainty is a limit of the sample rather than a finding about calibration.

\begin{table}[H]
\centering
\small
\caption{CCC with and without calibration}
\begin{tabular}{lrrrr}
\toprule
Factor & Uncalibrated & Calibrated & Gain & Gain 95\% CI \\
\midrule
$\Phi$\textsubscript{II} & \phantom{$-$}0.251 & \phantom{$-$}0.293 & \phantom{$-$}0.042 & $-$0.062 to 0.159 \\
$\Phi$\textsubscript{NPQ} & \phantom{$-$}0.127 & \phantom{$-$}\textbf{0.293} & \phantom{$-$}0.166 & \phantom{$-$}0.043 to 0.304 \\
$\Phi$\textsubscript{NO} & \phantom{$-$}0.108 & \phantom{$-$}\textbf{0.393} & \phantom{$-$}0.285 & \phantom{$-$}0.158 to 0.393 \\
Fv$'$/Fm$'$ & \phantom{$-$}0.079 & \phantom{$-$}\textbf{0.311} & \phantom{$-$}0.232 & \phantom{$-$}0.089 to 0.385 \\
LEF & \phantom{$-$}0.453 & \phantom{$-$}\textbf{0.554} & \phantom{$-$}0.101 & \phantom{$-$}0.035 to 0.168 \\
LTD & \phantom{$-$}0.008 & \phantom{$-$}\textbf{0.234} & \phantom{$-$}0.226 & \phantom{$-$}0.045 to 0.388 \\
CWSI & \phantom{$-$}0.019 & \phantom{$-$}0.198 & \phantom{$-$}0.179 & $-$0.024 to 0.380 \\
\bottomrule
\end{tabular}
\end{table}

An uncalibrated species loses position before it loses movement. Uncalibrated, CCC falls to 0.008 on LTD and 0.019 on CWSI against a no-skill value of zero, while trajectory r holds near 0.18 on both. Calibration requires labeled readings from the new species and a retrained head. Calibration does not require any change to the foundation model, which continues serving every readout built on it.

\subsection{A second sensor raises the performance floor without raising the ceiling}

Every plant in \S{}3.2's evaluation carries two sensors, so the value of a second sensor is measurable without a new experiment. The ``One sensor'' row below is \S{}3.2's result, unchanged.

\begin{table}[H]
\centering
\footnotesize
\setlength{\tabcolsep}{3pt}
\caption{Single-sensor and blended tracking (trajectory r)}
\begin{tabular}{lrrrrrrr}
\toprule
 & $\Phi$\textsubscript{II} & $\Phi$\textsubscript{NPQ} & $\Phi$\textsubscript{NO} & Fv$'$/Fm$'$ & LEF & LTD & CWSI \\
\midrule
Worse sensor & \phantom{$-$}0.258 & \phantom{$-$}0.312 & \phantom{$-$}0.319 & \phantom{$-$}0.312 & \phantom{$-$}0.620 & \phantom{$-$}0.134 & \phantom{$-$}0.106 \\
One sensor & \phantom{$-$}0.431 & \phantom{$-$}0.452 & \phantom{$-$}0.448 & \phantom{$-$}0.450 & \phantom{$-$}0.707 & \phantom{$-$}0.292 & \phantom{$-$}0.269 \\
\textbf{Two, blended} & \phantom{$-$}0.498 & \phantom{$-$}0.507 & \phantom{$-$}0.502 & \phantom{$-$}0.504 & \phantom{$-$}0.747 & \phantom{$-$}0.368 & \phantom{$-$}0.330 \\
Better sensor & \phantom{$-$}0.577 & \phantom{$-$}0.573 & \phantom{$-$}0.561 & \phantom{$-$}0.569 & \phantom{$-$}0.777 & \phantom{$-$}0.436 & \phantom{$-$}0.418 \\
\bottomrule
\end{tabular}
\end{table}

The two sensors on any given plant disagree by 0.157 to 0.319 in trajectory r, and on LTD and CWSI that gap is wider than the entire ``One sensor'' result. Blending both sensors (\S{}2.11) lands at 72 to 81 percent of the way up the range from the worse sensor to the better one, and beats the worse sensor on 94 to 100 percent of plants. The blend's gain over one sensor is 0.039 to 0.076, with every confidence interval clear of zero. The other measures also improve: CCC by 0.017 to 0.028, AUC by 0.021 to 0.033, and RMSE-s$\sigma$ by 0.039 to 0.071.

\begin{figure}[H]
\centering
\includegraphics{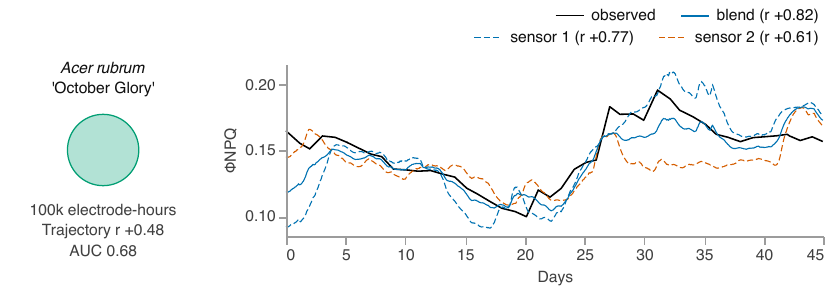}
\caption{$\Phi$\textsubscript{NPQ} over 45 days for one plant from the test split of \emph{Acer rubrum} `October Glory'. The two sensors on the plant and their blend are drawn against the observed ground-truth readings, each labeled with its trajectory r. The circle is the species' labeled corpus on the same area scale as \S{}2.2's figures, and the trajectory r and AUC beside it are the species'.}
\end{figure}

Averaged over plants, the blend falls 0.048 to 0.088 short of the better of the two sensors, and on most plants the better sensor alone would have done better, but which of the two is better is knowable only in hindsight. The better sensor marks the top of the range rather than an option a deployment could take. Blending stops short of that top because the two sensors share most of what limits them. The two sensors carry nearly the same per-sensor offset (\S{}2.7.4) on a plant, and across plants those offsets correlate at 0.913 to 0.954. Averaging two independent offsets would remove half their variance, but averaging the two sensors' near-identical offsets removes only a few percent of their variance.

Adding a second sensor cannot correct the error that belongs to the plant and to the leaf chosen to represent that plant. What a second sensor does correct is the error that differs between sensors, and that correction applies to nearly every plant.

\section{Discussion}

\subsection{A general representation of physiological state}

A plant's electrical activity carries information about its physiological state [Li et al. 2021]. Our results show that a self-supervised model can learn a general representation of the plant's state once and support a physiological readout (the stress-factor head) on top of it. All six of the results in \S{}3 rest on one backbone with two heads on it: the patch-reconstruction head of \S{}3.1 and the stress-factor head of \S{}3.2 through \S{}3.6.

Three of our findings measure the same property, generalizability across species, each by a different route. The foundation model captures the structure of the plant electrome in species held out of its pretraining (\S{}3.1). Its advantage grows when measured across more species (\S{}3.3). The stress-factor head tracks every factor on a species with no labels of its own (\S{}3.5). Taken together, these results show that the foundation model's representation carries across species.

\S{}3.3 and \S{}3.5 agree on a different property: widening the population improves performance on position more than movement, and an uncalibrated species degrades in position before movement. Position is more fragile than movement, and \S{}3.3 and \S{}3.5 both arrive at that conclusion.

\subsection{Where the foundation model matters most}

The improved performance that the foundation model provides is greater outside the six densely labeled species than inside them (\S{}3.3), so the denser a species' labels, the less pretraining adds. Therefore, the foundation model's value is greatest on the sparsely labeled species, and on the far larger number that carry no labels at all. Pretraining needs no labels of its own, so extending the foundation model's breadth requires only more electrical recordings, rather than more costly ground-truth labels.

\subsection{Non-drought stressors}

The head's estimates move in response to untrained stressors as we expect based on the existing literature (\S{}3.4), suggesting that it reads the plant's true physiological state rather than a signature specific to water stress.

The two stressors (salt and glyphosate) act through unrelated physiological mechanisms and the literature treats them separately. At a lethal glyphosate dose, $\Phi$\textsubscript{II} and dark-adapted Fv/Fm fall alongside CO\textsubscript{2} uptake over the days between application and death [Naidoo \& Naidoo 2025]. The head estimates the light-adapted Fv$'$/Fm$'$, which we match to that dark-adapted Fv/Fm. Fv$'$/Fm$'$ and Fv/Fm both track photosystem II efficiency and fall as that efficiency declines, so the expected direction transfers between them, though the values differ: the light-adapted Fv$'$/Fm$'$ is taken with non-photochemical quenching active, which lowers it. Severe salinity lowers $\Phi$\textsubscript{II} and Fv$'$/Fm$'$ while raising non-photochemical quenching [Huang et al. 2014]. Neither study gives a direction for $\Phi$\textsubscript{NO}, so we read it as the remaining share of the partition (\S{}2.4).

Our results (\S{}3.4) align with these expectations, but four constraints bound how far the results reach.

\begin{enumerate}

\item The DGP is underpowered to contrast its two stressors against each other, rather than against control.

\item The DGP was run on \emph{Coleus scutellarioides} while the literature from which predictions were derived is drawn from other species.

\item The salt expectation is extrapolated from sublethal work, while we administered a lethal dose.

\item The glyphosate literature reports the dark-adapted Fv/Fm rather than the light-adapted Fv$'$/Fm$'$ the head estimates.

\end{enumerate}

\subsection{Limitations}

This paper reports one physiological readout. Adding a second physiological readout is a matter of collecting a new kind of ground truth, not of changing the architecture (\S{}2.1).

\subsubsection{Species and ground-truth limitations}

The head's performance varies by species, and on \emph{Codiaeum variegatum} the head recovers nothing (\S{}3.2). \emph{Codiaeum variegatum} holds the smallest labeled corpus out of the six densely labeled species, but that species' evaluation rests on only four plants and eight sensors in the test split. Because 9 to 25 percent of sensors track negatively depending on the factor (\S{}3.2), the poor performance may be an anomaly of those particular plants and sensors rather than a general failure of the stress-factor head to read the species.

All of our ground-truth readings come from the MultispeQ, and our use of it has limits of its own. The MultispeQ's leaf temperature readings were not viable in \emph{Phoenix roebelenii} or \emph{Picea abies}, the only palm and the only conifer in the densely labeled set, so the two thermal stress factors are not reported for them. We have not isolated whether that reflects the instrument, the growth form, or those particular DGPs. Because each plant's MultispeQ readings come from one tagged leaf (\S{}2.4.2), a leaf's own offset stays constant for that plant, exactly as a model offset would. The data cannot separate the head's per-sensor offset from the leaf's offset, and the position measures (CCC, RMSE-s$\sigma$, and AUC) carry both offsets combined.

\subsubsection{Dataset limitations}

Plant responses are history-dependent, since a pre-stressed plant responds differently to a new stress than a naive plant [Pissolato et al. 2024; Oberkofler et al. 2021]. Our data comes from outdoor, greenhouse, and laboratory settings, but no single plant is recorded for longer than three months. Capturing a full lifecycle and its seasonal variation will require recordings that span more of a plant's life. Our evaluations do not include a site held out from training, so we cannot say how the readout transfers to a site at which the model has never recorded. Our evaluations also do not yet separate site from species, because no species is recorded at more than one site, so we cannot measure site differences and species differences separately.

The stress-factor head's labels come only from plants under drought and normal watering. Those two treatments set both the range of states the head can express and the conditions against which the head can be validated. Labeled recordings under salt, glyphosate, and other stressors would make \S{}3.4's directional arguments quantitative and more convincing, without the need for extrapolation from the literature.

\subsection{Future work}

Follow-on work proceeds in five directions.

\begin{enumerate}

\item Collecting more data, from more field sites, across seasons and full plant lifecycles, on more species the head has never been trained for, and under more stressors beyond drought.

\item Adding more physiological readouts to the model, each needing a new kind of ground truth and a new head.

\item Evolving the architecture and reading several sensors on one plant simultaneously, rather than blending the sensors' separate estimates post hoc.

\item Testing whether the structure within the minute, which per-minute aggregation obscures (\S{}2.3), would enrich the model's representation.

\item Closing the loop, putting plant care (e.g., irrigation) under Bombadil's automated control and collecting data from a system run in this fashion.

\end{enumerate}

Post-hoc blending of multiple sensors' estimates raises only the floor of the performance range over a single sensor (\S{}3.6). Raising the top of that range may require giving the model a fuller view of the plant's electrical state, from several sensors at once. Electrical activity synchronizes across different parts of a plant after a stimulus [Oliveira et al. 2025], and a model reading several sensors together could learn from that synchronization, where a blend of separate estimates cannot. That is one possible route to the higher tracking performance that the labels permit but the head does not yet reach.

\section{Conclusion}

Until now, a new factor, species, or condition has meant a new labeled dataset and a new model. Bombadil learns the structure of the plant electrome without labels, so one representation serves new factors, species, and conditions alike. Our first physiological head on Bombadil's representation estimates seven stress factors within a plant, reads species with no labels of their own, and moves as the literature predicts under stressors it never trained on. Bombadil can expand to further readouts needing only a new head, and to new species needing only electrical signal recordings rather than costly ground-truth labels. Continuous, species-agnostic measurement of a plant's physiological state from electrical signals alone is unlocked by Bombadil's understanding of the fundamental structure of the plant electrome.

\section*{Acknowledgements}

For their support, advice, encouragement, feedback, vision, and clarity, Electric Plant Co would like to express appreciation for Dave Clifford, Royal Heins, Vicki Boykis, Noah Langenfeld, Peter Sobot, Gustavo Souza, Michelle Gill of NVIDIA, Jeremy Jordan of NVIDIA, Tom McQuillen of Onto Ventures, Dan Fitzgerald of Onto Ventures, Elena Klonsky of Onto Ventures, and many others.

\refsbreak

\section*{References}

\begingroup\small\setlength{\parindent}{0pt}\raggedright

\hangindent=1.5em Ansari, A. F., Stella, L., T\"{u}rkmen, A. C., Zhang, X., Mercado, P., Shen, H., Shchur, O., Rangapuram, S. S., Pineda-Arango, S., Kapoor, S., Zschiegner, J., Maddix, D. C., Wang, H., Mahoney, M. W., Torkkola, K., Wilson, A. G., Bohlke-Schneider, M., \& Wang, Y. (2024). Chronos: Learning the language of time series. Transactions on Machine Learning Research. https://openreview.net/forum?id=gerNCVqqtR\par\vspace{3pt}

\hangindent=1.5em Baker, N. R. (2008). Chlorophyll fluorescence: A probe of photosynthesis in vivo. Annual Review of Plant Biology, 59(1), 89-113. https://doi.org/10.1146/annurev.arplant.59.032607.092759\par\vspace{3pt}

\hangindent=1.5em Bielczynski, L. W., \L{}\k{a}cki, M. K., Hoefnagels, I., Gambin, A., \& Croce, R. (2017). Leaf and plant age affects photosynthetic performance and photoprotective capacity. Plant Physiology, 175(4), 1634-1648. https://doi.org/10.1104/pp.17.00904\par\vspace{3pt}

\hangindent=1.5em Bommasani, R., Hudson, D. A., Adeli, E., Altman, R., Arora, S., von Arx, S., Bernstein, M. S., Bohg, J., Bosselut, A., Brunskill, E., Brynjolfsson, E., Buch, S., Card, D., Castellon, R., Chatterji, N., Chen, A., Creel, K., Davis, J. Q., Demszky, D., ... Liang, P. (2021). On the opportunities and risks of foundation models. arXiv. https://doi.org/10.48550/arXiv.2108.07258\par\vspace{3pt}

\hangindent=1.5em Chatterjee, S. K., Das, S., Maharatna, K., Masi, E., Santopolo, L., Mancuso, S., \& Vitaletti, A. (2015). Exploring strategies for classification of external stimuli using statistical features of the plant electrical response. Journal of the Royal Society Interface, 12(104), Article 20141225. https://doi.org/10.1098/rsif.2014.1225\par\vspace{3pt}

\hangindent=1.5em Cohen, J. (1988). Statistical power analysis for the behavioral sciences (2nd ed.). Lawrence Erlbaum Associates. ISBN 0-8058-0283-5.\par\vspace{3pt}

\hangindent=1.5em Comparini, D., Masi, E., Pandolfi, C., Sabbatini, L., Dolfi, M., Morosi, S., \& Mancuso, S. (2020). Stem electrical properties associated with water stress conditions in olive tree. Agricultural Water Management, 234, Article 106109. https://doi.org/10.1016/j.agwat.2020.106109\par\vspace{3pt}

\hangindent=1.5em Das, A., Kong, W., Sen, R., \& Zhou, Y. (2024). A decoder-only foundation model for time-series forecasting. In Proceedings of the 41st International Conference on Machine Learning (Proceedings of Machine Learning Research, Vol. 235, pp. 10148-10167). PMLR. https://proceedings.mlr.press/v235/das24c.html\par\vspace{3pt}

\hangindent=1.5em de Toledo, G. R. A., Parise, A. G., Simmi, F. Z., Costa, A. V. L., Senko, L. G. S., Debono, M.-W., \& Souza, G. M. (2019). Plant electrome: The electrical dimension of plant life. Theoretical and Experimental Plant Physiology, 31(1), 21-46. https://doi.org/10.1007/s40626-019-00145-x\par\vspace{3pt}

\hangindent=1.5em de Toledo, G. R. A., Reissig, G. N., Senko, L. G. S., Pereira, D. R., da Silva, A. F., \& Souza, G. M. (2024). Common bean under different water availability reveals classifiable stimuli-specific signatures in plant electrome. Plant Signaling \& Behavior, 19(1), Article 2333144. https://doi.org/10.1080/15592324.2024.2333144\par\vspace{3pt}

\hangindent=1.5em Devlin, J., Chang, M.-W., Lee, K., \& Toutanova, K. (2019). BERT: Pre-training of deep bidirectional transformers for language understanding. In Proceedings of the 2019 Conference of the North American Chapter of the Association for Computational Linguistics: Human Language Technologies, Volume 1 (Long and Short Papers) (pp. 4171-4186). Association for Computational Linguistics. https://doi.org/10.18653/v1/N19-1423\par\vspace{3pt}

\hangindent=1.5em Felle, H. H., \& Zimmermann, M. R. (2007). Systemic signalling in barley through action potentials. Planta, 226(1), 203-214. https://doi.org/10.1007/s00425-006-0458-y\par\vspace{3pt}

\hangindent=1.5em Fromm, J., \& Lautner, S. (2007). Electrical signals and their physiological significance in plants. Plant, Cell \& Environment, 30(3), 249-257. https://doi.org/10.1111/j.1365-3040.2006.01614.x\par\vspace{3pt}

\hangindent=1.5em Gil, P. M., \& Vargas, A. I. (2023). Stem electrical potential variations may aid in the early detection of drought stress in fruit-bearing trees. International Journal of Agriculture and Natural Resources, 50(3), 116-129. https://doi.org/10.7764/ijanr.v50i3.2552\par\vspace{3pt}

\hangindent=1.5em Gonz\'{a}lez i Jucl\`{a}, D., Najdenovska, E., Dutoit, F., \& Raileanu, L. E. (2023). Detecting stress caused by nitrogen deficit using deep learning techniques applied on plant electrophysiological data. Scientific Reports, 13, Article 9633. https://doi.org/10.1038/s41598-023-36683-3\par\vspace{3pt}

\hangindent=1.5em He, K., Chen, X., Xie, S., Li, Y., Doll\'{a}r, P., \& Girshick, R. (2022). Masked autoencoders are scalable vision learners. In Proceedings of the 2022 IEEE/CVF Conference on Computer Vision and Pattern Recognition (CVPR) (pp. 15979-15988). IEEE. https://doi.org/10.1109/CVPR52688.2022.01553\par\vspace{3pt}

\hangindent=1.5em Huang, C., Wei, G., Jie, Y., Wang, L., Zhou, H., Ran, C., Huang, Z., Jia, H., \& Anjum, S. A. (2014). Effects of concentrations of sodium chloride on photosynthesis, antioxidative enzymes, growth and fiber yield of hybrid ramie. Plant Physiology and Biochemistry, 76, 86-93. https://doi.org/10.1016/j.plaphy.2013.12.021\par\vspace{3pt}

\hangindent=1.5em Huber, A. E., \& Bauerle, T. L. (2016). Long-distance plant signaling pathways in response to multiple stressors: The gap in knowledge. Journal of Experimental Botany, 67(7), 2063-2079. https://doi.org/10.1093/jxb/erw099\par\vspace{3pt}

\hangindent=1.5em Idso, S. B., Jackson, R. D., Pinter, P. J., Jr., Reginato, R. J., \& Hatfield, J. L. (1981). Normalizing the stress-degree-day parameter for environmental variability. Agricultural Meteorology, 24, 45-55. https://doi.org/10.1016/0002-1571(81)90032-7\par\vspace{3pt}

\hangindent=1.5em Jackson, R. D., Idso, S. B., Reginato, R. J., \& Pinter, P. J., Jr. (1981). Canopy temperature as a crop water stress indicator. Water Resources Research, 17(4), 1133-1138. https://doi.org/10.1029/WR017i004p01133\par\vspace{3pt}

\hangindent=1.5em Jiang, W.-B., Zhao, L.-M., \& Lu, B.-L. (2024). Large brain model for learning generic representations with tremendous EEG data in BCI. In The Twelfth International Conference on Learning Representations (ICLR 2024). https://openreview.net/forum?id=QzTpTRVtrP\par\vspace{3pt}

\hangindent=1.5em Kostas, D., Aroca-Ouellette, S., \& Rudzicz, F. (2021). BENDR: Using transformers and a contrastive self-supervised learning task to learn from massive amounts of EEG data. Frontiers in Human Neuroscience, 15, Article 653659. https://doi.org/10.3389/fnhum.2021.653659\par\vspace{3pt}

\hangindent=1.5em Kramer, D. M., Johnson, G., Kiirats, O., \& Edwards, G. E. (2004). New fluorescence parameters for the determination of QA redox state and excitation energy fluxes. Photosynthesis Research, 79(2), 209-218. https://doi.org/10.1023/B:PRES.0000015391.99477.0d\par\vspace{3pt}

\hangindent=1.5em Kuhlgert, S., Austic, G., Zegarac, R., Osei-Bonsu, I., Hoh, D., Chilvers, M. I., Roth, M. G., Bi, K., TerAvest, D., Weebadde, P., \& Kramer, D. M. (2016). MultispeQ Beta: A tool for large-scale plant phenotyping connected to the open PhotosynQ network. Royal Society Open Science, 3(10), Article 160592. https://doi.org/10.1098/rsos.160592\par\vspace{3pt}

\hangindent=1.5em Lee, J., \& Calvo, P. (2023). The potential of plant action potentials. Synthese, 202(6), Article 176. https://doi.org/10.1007/s11229-023-04398-7\par\vspace{3pt}

\hangindent=1.5em Li, J.-H., Fan, L.-F., Zhao, D.-J., Zhou, Q., Yao, J.-P., Wang, Z.-Y., \& Huang, L. (2021). Plant electrical signals: A multidisciplinary challenge. Journal of Plant Physiology, 261, Article 153418. https://doi.org/10.1016/j.jplph.2021.153418\par\vspace{3pt}

\hangindent=1.5em Lin, L. I.-K. (1989). A concordance correlation coefficient to evaluate reproducibility. Biometrics, 45(1), 255-268. https://doi.org/10.2307/2532051 [Correction: Lin, L. I.-K. (2000). A note on the concordance correlation coefficient. Biometrics, 56(1), 324-325. https://doi.org/10.1111/j.0006-341x.2000.00324.x]\par\vspace{3pt}

\hangindent=1.5em Liu, G., Gao, X., Dresler, M., Zhang, J., \& Wei, P. (2026). Rethinking random Transformers as adaptive sequence smoothers for sleep staging. arXiv. https://doi.org/10.48550/arXiv.2605.09905\par\vspace{3pt}

\hangindent=1.5em Mudrilov, M., Ladeynova, M., Grinberg, M., Balalaeva, I., \& Vodeneev, V. (2021). Electrical signaling of plants under abiotic stressors: Transmission of stimulus-specific information. International Journal of Molecular Sciences, 22(19), Article 10715. https://doi.org/10.3390/ijms221910715\par\vspace{3pt}

\hangindent=1.5em Naidoo, K. K., \& Naidoo, G. (2025). Photosynthetic response of the invasive weed \emph{Chromolaena odorata} to glyphosate. Photosynthesis Research, 163(6), Article 55. https://doi.org/10.1007/s11120-025-01177-y\par\vspace{3pt}

\hangindent=1.5em Najdenovska, E., Dutoit, F., Tran, D., Rochat, A., Vu, B., Mazza, M., Camps, C., Plummer, C., Wallbridge, N., \& Raileanu, L. E. (2021). Identifying general stress in commercial tomatoes based on machine learning applied to plant electrophysiology. Applied Sciences, 11(12), Article 5640. https://doi.org/10.3390/app11125640\par\vspace{3pt}

\hangindent=1.5em Nie, Y., Nguyen, N. H., Sinthong, P., \& Kalagnanam, J. (2023). A time series is worth 64 words: Long-term forecasting with transformers. In The Eleventh International Conference on Learning Representations (ICLR 2023). https://arxiv.org/abs/2211.14730\par\vspace{3pt}

\hangindent=1.5em Oberkofler, V., Pratx, L., \& B\"{a}urle, I. (2021). Epigenetic regulation of abiotic stress memory: Maintaining the good things while they last. Current Opinion in Plant Biology, 61, Article 102007. https://doi.org/10.1016/j.pbi.2021.102007\par\vspace{3pt}

\hangindent=1.5em Oliveira, T. F. de C., Parise, A. G., Tasca, H. C., Posso, D., Reissig, G. N., \& Souza, G. M. (2025). Abiotic stress triggers electrical synchronisation of shoot and leaves in soybean plants: A clue for plant attention-like. Plant Signaling \& Behavior, 20(1), Article 2577400. https://doi.org/10.1080/15592324.2025.2577400\par\vspace{3pt}

\hangindent=1.5em Pissolato, M. D., Martins, T. S., Fajardo, Y. C. G., Souza, G. M., Machado, E. C., \& Ribeiro, R. V. (2024). Stress memory in crops: What we have learned so far. Theoretical and Experimental Plant Physiology, 36(3), 535-565. https://doi.org/10.1007/s40626-024-00315-6\par\vspace{3pt}

\hangindent=1.5em Qi, J., Liu, C., Wang, Q., Shi, Y., Xia, X., Wang, H., Sun, L., \& Men, H. (2024). Clivia biosensor: Soil moisture identification based on electrophysiology signals with deep learning. Biosensors and Bioelectronics, 262, Article 116525. https://doi.org/10.1016/j.bios.2024.116525\par\vspace{3pt}

\hangindent=1.5em Qiu, W., Huang, Z., Hu, H., Feng, A., Yan, Y., \& Ying, Z. (2025). MindLLM: A subject-agnostic and versatile model for fMRI-to-text decoding. In Proceedings of the 42nd International Conference on Machine Learning (Proceedings of Machine Learning Research, Vol. 267, pp. 50572-50593). PMLR. https://proceedings.mlr.press/v267/qiu25e.html\par\vspace{3pt}

\hangindent=1.5em R\'{i}os-Rojas, L., Tapia, F., \& Gurovich, L. A. (2014). Electrophysiological assessment of water stress in fruit-bearing woody plants. Journal of Plant Physiology, 171(10), 799-806. https://doi.org/10.1016/j.jplph.2014.02.005\par\vspace{3pt}

\hangindent=1.5em Sade, N., Gebremedhin, A., \& Moshelion, M. (2012). Risk-taking plants: Anisohydric behavior as a stress-resistance trait. Plant Signaling \& Behavior, 7(7), 767-770. https://doi.org/10.4161/psb.20505\par\vspace{3pt}

\hangindent=1.5em Saxe, A. M., Koh, P. W., Chen, Z., Bhand, M., Suresh, B., \& Ng, A. Y. (2011). On random weights and unsupervised feature learning. In Proceedings of the 28th International Conference on Machine Learning (ICML 2011) (pp. 1089-1096). Omnipress. https://icml.cc/2011/papers/551\_icmlpaper.pdf\par\vspace{3pt}

\hangindent=1.5em Sperdouli, I., Moustaka, J., Ouzounidou, G., \& Moustakas, M. (2021). Leaf age-dependent photosystem II photochemistry and oxidative stress responses to drought stress in \emph{Arabidopsis thaliana} are modulated by flavonoid accumulation. Molecules, 26(14), Article 4157. https://doi.org/10.3390/molecules26144157\par\vspace{3pt}

\hangindent=1.5em Stahlberg, R., Cleland, R. E., \& Van Volkenburgh, E. (2006). Slow wave potentials --- a propagating electrical signal unique to higher plants. In F. Balu\v{s}ka, S. Mancuso, \& D. Volkmann (Eds.), Communication in plants: Neuronal aspects of plant life (pp. 291-308). Springer. https://doi.org/10.1007/978-3-540-28516-8\_20\par\vspace{3pt}

\hangindent=1.5em Sukhov, V., Sukhova, E., \& Vodeneev, V. (2019). Long-distance electrical signals as a link between the local action of stressors and the systemic physiological responses in higher plants. Progress in Biophysics and Molecular Biology, 146, 63-84. https://doi.org/10.1016/j.pbiomolbio.2018.11.009\par\vspace{3pt}

\hangindent=1.5em Tran, D., Dutoit, F., Najdenovska, E., Wallbridge, N., Plummer, C., Mazza, M., Raileanu, L. E., \& Camps, C. (2019). Electrophysiological assessment of plant status outside a Faraday cage using supervised machine learning. Scientific Reports, 9, Article 17073. https://doi.org/10.1038/s41598-019-53675-4\par\vspace{3pt}

\hangindent=1.5em Tran, D., Najdenovska, E., Dutoit, F., Plummer, C., Wallbridge, N., Mazza, M., Camps, C., \& Raileanu, L. E. (2024). Advanced assessment of nutrient deficiencies in greenhouse with electrophysiological signals. Horticulture, Environment, and Biotechnology, 65(4), 567-580. https://doi.org/10.1007/s13580-023-00589-w\par\vspace{3pt}

\hangindent=1.5em Vaswani, A., Shazeer, N., Parmar, N., Uszkoreit, J., Jones, L., Gomez, A. N., Kaiser, \L{}., \& Polosukhin, I. (2017). Attention is all you need. In Advances in Neural Information Processing Systems 30 (pp. 5998-6008). https://arxiv.org/abs/1706.03762\par\vspace{3pt}

\hangindent=1.5em Wahid, A. M., Baihaqi, W. M., \& Nambo, H. (2026). Transformer adaptation for plant bioelectric potential classification: A scoping review and analogical transfer framework. Smart Agricultural Technology, 15, Article 102484. https://doi.org/10.1016/j.atech.2026.102484\par\vspace{3pt}

\hangindent=1.5em Zerveas, G., Jayaraman, S., Patel, D., Bhamidipaty, A., \& Eickhoff, C. (2021). A Transformer-based framework for multivariate time series representation learning. In Proceedings of the 27th ACM SIGKDD Conference on Knowledge Discovery \& Data Mining (pp. 2114-2124). Association for Computing Machinery. https://doi.org/10.1145/3447548.3467401\par\vspace{3pt}

\hangindent=1.5em Zimmermann, M. R., Maischak, H., Mith\"{o}fer, A., Boland, W., \& Felle, H. H. (2009). System potentials, a novel electrical long-distance apoplastic signal in plants, induced by wounding. Plant Physiology, 149(3), 1593-1600. https://doi.org/10.1104/pp.108.133884\par\vspace{3pt}

\endgroup
\end{document}